\documentclass{article} 
\usepackage{iclr2022_conference,times}

\usepackage{amsmath,amsfonts,bm}

\def\eqref#1{equation~\ref{#1}}

\def\1{\bm{1}}

\DeclareMathAlphabet{\mathsfit}{\encodingdefault}{\sfdefault}{m}{sl}
\SetMathAlphabet{\mathsfit}{bold}{\encodingdefault}{\sfdefault}{bx}{n}

\usepackage{tabularx}
\usepackage[utf8]{inputenc} 
\usepackage[T1]{fontenc}    
\usepackage{hyperref}       
\usepackage{url}            
\usepackage{booktabs}       
\usepackage{amsfonts}       
\usepackage{nicefrac}       
\usepackage{microtype}      
\usepackage{times}
\usepackage{helvet}
\usepackage{courier}
\usepackage{graphicx}
\usepackage{colortbl}

\title{Neural Logic Analogy Learning}

\iclrfinalcopy

\author{Yujia Fan  \\
Department of Computer Science\\
Rutgers University\\
New Brunswick, NJ, USA \\
\texttt{yujia.fan@rutgers.edu} \\
\And
Yongfeng Zhang\\
Department of Computer Science \\
Rutgers University \\
New Brunswick, NJ, USA \\
\texttt{yongfeng.zhang@rutgers.edu} \\
}

\begin{document}

\maketitle

\begin{abstract}
Letter-string analogy is an important analogy learning task which seems to be easy for humans but very challenging for machines.
The main idea behind current approaches to solving letter-string analogies is to design heuristic rules for extracting analogy structures and constructing analogy mappings. However, one key problem is that it is difficult to build a comprehensive and exhaustive set of analogy structures which can fully describe the subtlety of analogies. This problem makes current approaches unable to handle complicated letter-string analogy problems.
In this paper, we propose \textbf{N}eural l\textbf{o}gic \textbf{a}nalogy lear\textbf{n}ing (Noan), which is a dynamic neural architecture driven by differentiable logic reasoning to solve analogy problems. Each analogy problem is converted into logical expressions consisting of logical variables and basic logical operations (AND, OR, and NOT). More specifically, Noan learns the logical variables as vector embeddings and learns each logical operation as a neural module. In this way, the model builds computational graph integrating neural network with logical reasoning to capture the internal logical structure of the input letter strings. The analogy learning problem then becomes a True/False evaluation problem of the logical expressions. 
Experiments show that our machine learning-based Noan approach outperforms state-of-the-art approaches on standard letter-string analogy benchmark datasets. 
\end{abstract}

\section{Introduction}
As an engine of cognition, analogy plays an important role in categorization, decision making, problem solving, and creative discovery \citep{gentner2012analogical}.
An analogy is a comparison between two objects, or systems of objects, that highlights respects in which they are thought to be similar \citep{bartha2013analogy}. 
Letter-string analogy is a type of analogy that can be written in the following form $a:b::c:d$, meaning that $a$ is to $b$ what $c$ is to $d$, where $a$, $b$, $c$, $d$ are letter strings, $a$ is the initial string, $b$ is the modified string, $c$ is the query string, and $d$ is the answer string.
For example, ``ABC:ABD::IJK:IJL'' is a specific letter-string analogy, which means that if ABC changes to ABD, then analogously IJK should change to IJL. 
A letter-string analogy question is usually asked in the following way: ``ABC:ABD::IJK:?'' which reads if ABC changes to ABD, then how should IJK change in an analogous way? Here ``ABC:ABD'' is the given background knowledge, IJK is the query string, and the question asks for the correct answer string. More complicated analogy questions could be ``ABAC:ACAB::DEFG:?'' and a good answer would be DGFE since the analogy is switching the second and fourth letter. 

Though seems to be relatively easy for humans, analogy learning is difficult for machines for three reasons. First, one key challenge is that there could be various different types of analogous relations, and thus it is very difficult to manually design universal rules or models for analogy learning. Second, many analogy problems include letter manipulation in a discrete space (as shown in the above examples), which makes it difficult to train differentiable machine learning models in continuous space.
Finally, designing models for analogy learning not only needs perceptual learning and pattern recognition from data but also certain degree of cognitive reasoning ability.
As a result, the letter-string analogy learning problem is an ideal laboratory to study human's high-level perception since it actually shows remarkable degree of subtlety \citep{marshall1997metacat}.
Several computational models have been proposed to solve letter-string analogies.
For example, Copycat \citep{hofstadter1994copycat} and its successor Metacat \citep{marshall1997metacat} developed by Hofstadter et al.  characterize the transformation process of the initial string, and construct mappings between the initial string and the target string to generate answers.
Murena \citep{murena2017complexity} developed a new generative language to describe analogy problems and proposed that the optimal solution of an analogy problem has minimum complexity \citep{li2008introduction}.
\cite{rijsdijksolving} solved letter-string analogies based on the hybrid inferential process integrating structural information theory, which is a framework used to predict phenomena of perceptual organization based on complexity metrics.
However, there exist weaknesses in these approaches in terms of describing analogy problems, building transformation structure between initial string and modified string, and constructing mapping between initial string and target string. 

Recently, Neural Logic Reasoning has become a promising approach to integrating neural network learning and cognitive reasoning \citep{shi2020neural,chen2021neural,chen2022graph}. This paper proposes \textbf{N}eural l\textbf{o}gic \textbf{a}nalogy lear\textbf{n}ing (Noan), a dynamic neural architecture to solve analogies based on Logic-Integrated Neural Networks (LINN) \citep{shi2020neural}. We convert each analogy problem to a logical expression consisting of logical variables and basic logical operations such as AND, OR, and NOT. Noan regards the logical variables as vector embeddings and adopts each basic operation as a neural module based on logical regularization. The model then builds computational graph to integrate neural network with logic reasoning to capture the structure information of the analogy expressions. The analogy problem then becomes a True/False evaluation problem of the logical expressions. 
Since Noan is based on differentiable machine learning rather than designing discrete mapping structures, the weaknesses in previous approaches mentioned above can be largely avoided.
Furthermore, experiments on benchmark letter-string analogy datasets show the superior performance of our approach compared with structure mapping and complexity computing approaches. 

To the best of our knowledge, this is one of the first work to apply machine learning based model to solve letter-string analogies. In the following, we explain the details of our proposed model in Section \ref{sec:model}, compare with several baseline models through two analogy datasets in Section \ref{sec:experiment}, and conclude the work together with future research directions in Section \ref{sec:conclusion}. Related work and a review of neural logic reasoning are presented in Appendix \ref{sec:related_work} and \ref{sec:preliminary}, respectively.

\section{Neural Logic Analogy Learning}
\label{sec:model}


\subsection{Reasoning with Commonsense Data and One-shot Data}
One fundamental requirement of solving analogies for human beings is to start from commonsense data and reason towards correct answers. For example, human will rely on their basic commonsense to solve problems, i.e., every participant knows a total of 26 letters and should be familiar with their positional relationship. This inspires us to improve the inner-workings of pre-trained data for commonsense reasoning. We will first consider basic commonsense data, i.e., human level understanding about the alphabetical order. Consider the simplest one-character situation, the problem $A\rightarrow A$ is supposed to be true in commonsense, which will be the same case for the two-character situation: $AA\rightarrow AA$. And also, human level analogy understanding relies on specific order between letters. In human cognition, $A$ is followed by $B$, and $B$ is followed by $C$. So we can infer that $A\rightarrow B$, and $B\rightarrow C$. In order for our model to learn the strict sequence of the letters, we only allow derivation under adjacent neighbors. That means $A\rightarrow D$ is considered false in our commonsense dataset.
We only choose one-order and two-order training data to train the Noan model and it is expected to be sufficient to derive higher-order data. We include three fundamental logical relationships in this commonsense dataset: repetition, forward derivation, and reverse derivation. Accordingly, they follow the pattern of $A\rightarrow A$, $A\rightarrow B$ and $B\rightarrow A$ as well as $AA\rightarrow AA$, $AB\rightarrow BC$ and $BC\rightarrow AB$. Furthermore, any event that does not belong to these positive events mentioned above is considered as negative event. In our design, we randomly choose as many negative events as positive events to make sure the negation module can be also adequately developed in the training process. 

In addition to the commonsense data, the model is provided an analogy question such as $ABC : ABD :: IJK : ?$ to solve analogy problems, which means if $ABC$ transforms to $ABD$, then what should $IJK$ transform to.
In particular, $ABC : ABD$ encourages the use of  prior knowledge and raises the improvement on the original commonsense basis. In our model, this is called one-shot data.
Typically, one-shot data pretend to have more complicated inner logic which users can rely on to make a decision. Although it is just one piece of data, it plays a more important role in recognizing the pattern of the analogy. For instance, suppose a participant is provided with a single analogy problem without one-shot data: $III : ?$. Only depending on commonsense dataset, it is likely for us to give answers like $III$, $JJJ$, or $KKK$ which are all reasonable based on the possible logical guesses. However, different one-shot data could lead to totally divergent directions. If the given one-shot data is $AAA : A$, then the correct answer must be $III: I$ which is unpredictable only using commonsense dataset. In this way, one-shot data works as an anchor data point that defines the right orientation in the process of reasoning.
Based on the commonsense data and one-shot data, we can then assemble a neural architecture for the whole analogy problem. And it is worth noting that since the contents of one-shot data vary for different analogies, the structure and length of the logical expressions may also vary from each other, which would be dynamically assembled depending on different inputs.

\subsection{Neural Modules}
To transform each analogy statement into the neural logic expression, we first connect the letters in each sentence together by conjunction. In the analogy space, we totally have 26 variables $V={v_x}$, where $x\in \{A, B, ..., Z\}$. For example, $ABC$ would be interpreted as $v_A\land v_B \land v_C$. This is inline with the general perception since these three characters appear at the same time. And then we turn each analogy topic to the problem of deciding if the transformed implication statement is True or False, for example, a general logic expression
$A\land B \land C \rightarrow A\land B \land D$
can be written as
$v_A\land v_B \land v_C \rightarrow v_A\land v_B \land v_D$.
According to the material implication, this expression can be reinterpreted as
$\lnot (v_A\land v_B \land v_C) \lor (v_A\land v_B \land v_D)$.
This can be further reinterpreted into a simpler statement according to De Morgan's Law:
$ (\lnot v_A \lor \lnot v_B \lor \lnot v_C ) \lor (v_A\land v_B \land v_D)$.

In this way our model can turn literally logic statements into unique and quantitative forms. And then, to evaluate the True/False value of each expression, we evaluate the similarity between the expression vector and True vector. Here, \textbf{T} and \textbf{F} are True/False vector representations. In our Noan model, the module $Sim(\cdot, \cdot)$ is designed to calculate the similarity between two vectors and the output from $Sim(\cdot, \cdot)$ is expected to be in the range from 0 to 1. Necessarily, we define $E = \{e_i\}_{i=1}^m$ as a set of expressions and $Y = \{y_i\}_{i=1}^m$ as their according True/False values. And the similarity $p = Sim(\textbf{e}, \textbf{T})$ can be considered as the possibility that the expression is proven to be true. In our model, the similarity module is formulated as the cosine similarity between two vectors. We multiply the cosine similarity by a value $\alpha$, followed by a $sigmoid$ function:
$
Sim(\textbf{w}_i, \textbf{w}_j) = 
sigmoid \left(\alpha \frac{\textbf{w}_i \cdot \textbf{w}_j}{||\textbf{w}_i|| ||\textbf{w}_j||} \right)
$.
Here, \textbf{w} can be considered as a single vector or an expression in process of the neural modules and $\alpha$ is set to 10 to ensure the final output is formatted between 0 and 1 in the practical experiments.
To involve this output $p$ in the background of analogy solving, we consider the behavior of our Noan model to predict True/False values as a classification problem. And we choose the cross-entropy loss function as:
$ L_{loss} = - \sum_{e_i \in E} y_i log(p_i) + (1-y_i) log(1-p_i) $.

\subsection{Logical Regularization Neural Modules}
\begin{figure*}[h!]
\centering
\includegraphics[scale=0.28]{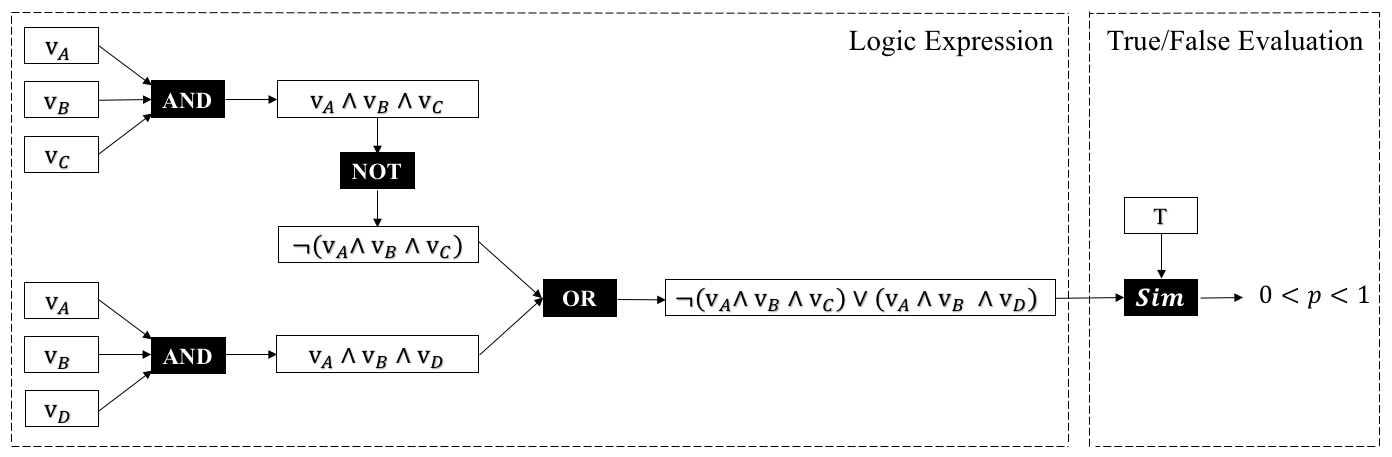}
\caption{An example of Noan.}
\label{fig:example}
\end{figure*}

So far, we have learnt three logical neural modules AND, OR, NOT as plain neural networks. However, not only should these neural modules perform the above three logic operations, we also need to guarantee they are really implementing the expected logic rules. For example, a double negation returns itself, $\lnot \lnot \textbf{w} = \textbf{w}$. To further apply such constraints to regularize the learning of the compound logic operations, we add logical regularizers to the previous neural modules, so that they will conduct certain logical rules. An entire set of these logical regularizers and their corresponding laws are listed in Table \ref{tab:logicalrules}.
In Table \ref{tab:logicalrules}, we translate these logical laws into equations represented by variables and modules in Noan. It should be noted that the vector space in Noan is not the whole vector space $\Bbb{R}^d$. Take Figure \ref{fig:example} as an example, the input variables like $\textbf{v}_A$, $\textbf{v}_B$, $\textbf{v}_C$, $\textbf{v}_D$, the intermediate expressions like $\textbf{v}_A \land \textbf{v}_B \land \textbf{v}_D$, $\lnot (\textbf{v}_A \land \textbf{v}_B \land \textbf{v}_C)$ and the final expressions like $ (\lnot \textbf{v}_A \lor \lnot \textbf{v}_B \lor \lnot \textbf{v}_C ) \lor (\textbf{v}_A\land \textbf{v}_B \land \textbf{v}_D)$ construct the vector space in Noan, which will be much smaller than the whole vector space $\Bbb{R}^d$. And also all above input variables as well as intermediate and final expressions are constrained by logical regularizers.
In Noan, we randomly generate the true vector \textbf{T} at the beginning and keep it fixed during the process of the training and testing. The true vector plays a anchor vector role in the whole space and accordingly, the false vector \textbf{F} is set as $\lnot \textbf{T}$. The result vector will be compared with the true vector \textbf{T} to decide the True/False output for each expression. 
Then, we combine the logical regularizers with the loss functions $L_{loss}$ defined before with weight $\lambda_l$.
Since a potential problem about the logical regularizers is that the vector length of logical variables or expressions may explode during the optimizing process of $L_1$, we add a common $\ell_2$-norm regularizer to the original loss function with weight $\lambda_\ell$ to limit the length of vectors to make the expected performance more stable.
Lastly, we add another $\ell_2$-length regularizer with weight $\lambda_\Theta$ to prevent the number of parameters from exploding. 
The final loss function which can prevent overfitting will be:
$ L = L_{loss} + \lambda_l \sum_{i} r_i + \lambda_{\ell} \sum_{w\in W} ||\textbf{w}||_F^2 + \lambda_{\Theta}||\Theta||_F^2 $,
where $r_i$ are the logical regularizers stated in Table \ref{tab:logicalrules}; $W$ include all set of input variables, intermediate and final expressions; $\Theta$ is the parameter group in the model.

\subsection{Model Prediction}
Our prototype model prediction is defined in this way: given a set of commonsense data and an one-shot data and their corresponding True/False values, we train a Noan model on a number of possible answers, and then predict the value of each expressions in the answer set and finally get the rank of these solutions. Since a possibility $p$ which returned by Noan model as an output falls between 0 and 1, it could be considered as a ranking criteria among those possible answers. Typically, the closer the value is to 1, the higher its ranking.
Theoretically, the number of possible answers is infinite but we would constrain the size of the answer set and manually give 20 most likely answers. For instance, to solve the analogy problem $AAABBB:AB::III:?$, we would explore an answer list of $I$, $II$, $III$, $J$, $IJ$, $IJK$, etc. Similar to the way we generate commonsense data, we consider three basic logical relationships: repetition like $I$, $II$, forward derivation like $J$, $IJ$, $IJK$, and reverse derivation like $JI$, $KJI$. To prevent artificial bias, we also include some randomly generated solutions of different lengths. 
To conclude, we conduct experiments on provided analogy expressions with the commonsense and one-shot data as the training data, the human-made answer set as the test data. To generate one part of  validation data, we follow the same pattern of the previous one-shot data to propagate as many expressions as possible. Given an one-shot data as $AAABBB:AB$, it is safe to derive $BBBCCC:BC$, $CCCDDD:CD$, etc. The other part of the validation data comes from commonsense data, which will further guarantee both of the datasets are adequately utilized during the training procedure.

\section{Experiments}
\label{sec:experiment}

\subsection{Datasets}

As the key motivation of this work is to develop a neural logic reasoning framework to solve letter-string analogy problems in a cognitively plausible manner, we prove the learning ability of Noan model to solve a variety of problems, including some that are previously unsolvable by cognition theory. We experiment with two publicly available datasets, Murena's dataset \citep{murena2017complexity} and Rijsdijk's dataset \citep{rijsdijksolving} with respect to real human-made answers. 
Murena's Dataset is conducted by \citeauthor{murena2017complexity} on human answers for analogy tests. Given the same template ABC:ABD::X:?. 68 participants were invited to solve the analogies with different X as shown in the first column of Table~\ref{tab:dataset1}. Two most selected answers, as well as the percentages of participants who choose these answers are presented in the second and the third column of Table~\ref{tab:dataset1}, respectively.
Rijsdijk's Dataset is a more complex dataset constructed by \citeauthor{rijsdijksolving}, which consists of 20 more complex analogies with various formats and patterns as shown in the first column in Table~\ref{tab:dataset2}. 
The second column of Table~\ref{tab:dataset2} presents the top two answers provided by 35 participants,
along with the percentages of the participants choosing these answers. Since all participants might offer the same answer or each participants gave different answers for the second top answer, only top answer is shown in some cases. 

To examine the effectiveness of the proposed neural logical reasoning model, we compare the performances with two other analogy making models, Metacat and Pisa (Parameter Load Plus ISA-rules). 
For all models, we provide an answer set consisting of 20 possible strings to the problem and ask the model to rank these strings. 
The last three columns of Table~\ref{tab:dataset1} and Table~\ref{tab:dataset2} show the performances of our Noan model($P_n$), Pisa($P_p$) and Metacat($P_m$) on the analogy solving of Murena's and Rijsdijk's datasets, in terms of the ranking in the given or the generated answers (e.g. 1 means the top 1 answer, 2 means the top 2 answer and so on). Besides, symbol $\infty$ shows that the top participant answer is not obtained by the approach.

\subsection{Experimental Results}

For Murena's dataset, overall, the top answer matches the most common participant answer 8/11 times (72.7$\%$) for all three approaches. The top 2 chosen or generated answers include the most common participant answer 11/11 times (100$\%$) for Noan and 10/11 times (90.9$\%$) for both Pisa and Metacat. Murena's dataset mainly considers the case where letters are moved forward. Similar performances on this dataset were obtained from the three approaches since these problems have the same format and pattern which all three algorithms can solve easily. 
Remarkably, we highlight three analogy problems $ABC:ABD::IJJKKK:?$, $ABC:ABD::RSSTTT:?$, and $ABC:ABD::MRRJJJ:?$ in Table \ref{tab:dataset1}. These three problems have same format, but the solutions for those three problems has different patterns, the first two problems are solved by changing all last three duplicated letters while the third problem is addressed through changing only the last one of the three duplicated letters. In the last problems, Noan fails to catch this kind of abnormal answer given by participants. This shows the subtleties of human thinking that humans sometimes will change their way of thinking according to different letters. However, the selection rates of the top two participant solutions are close, which means Noan still has a reasonable performance.

For Rijsdijk's dataset, overall, the top answer given by Noan was in the top two participant answers 19/20 times (95$\%$), whereas the top answer generated by Pisa and Metacat was in the top two participant answers 13/20 times (65$\%$) and 8/20 times (40$\%$), respectively. The most common participant answer matched the top generated 18/20 times (90$\%$) for Noan, 11/20 times (55$\%$) for Pisa, and 6/20 times (30$\%$) for Metacat. For this more complex dataset, our neural logic model offers more reasonable results compared to Pisa and Metacat. 
We noticed that it is rare that our model Noan misses the top answer and we highlighted these questions in red in Table \ref{tab:dataset2}. For example, the most common chosen solution for the problem $ABAC:ADAE::BACA:?$ is $DAEA$ which means the structure transformation between the initial string $ABAC$ and the target string $BACA$ (position swap of the first two letters and the last two letters) is more apparent than the transformation between the initial string $ABAC$ and the modified string $BACA$ (letter changes on specific positions). Noan gives priority to the latter structure transformation and regards $BCCC$ as the best solution; Pisa gives priority to the former structure transformation but offers the solution $BCCC$ a very low rank; Metacat is unable to handle this question. For this problem, our model Noan still gives the most reasonable answer.
There are more cases that Noan significantly outperforms other methods where Noan can give exact top answer while the other two method are even unable to solve it. And we highlighted them in green in Table \ref{tab:dataset2}. This performance shows that Noan is more capable to recognize swaps and duplicates. Specifically, for problems $ABC:BAC::IJKL:?$, $ABCD:CDAB::IJKLMN:?$, and $ABBA:BAAB::IJKL:?$, the most common participant solutions $JIKL$, $LMNIJK$, and $JILK$ are not obtained by Pisa and Metacat at the top rank but are provided by Noan. The key of those three problems is to swap letter positions. Specifically, $ABC:BAC$ shows the swap of the first two letters; $ABCD:CDAB$ presents the swap between the former two letters and the latter two letters; and $ABBA:BAAB$ means the first two letters swap and the last two letters swap, too. This shows Noan's stronger recognition ability for position exchange.
When it comes to the two similar problems $ABC:AAABBBCCC::ABCD:?$ and $ABC:ABBCCC::ABCD:?$, all three methods get the most common participant solution $AAABBBCCCDDD$ for the former analogy problem. However, only Noan obtains the top 1 common participant solution $ABBCCCDDDD$ for the latter analogy problem. Since the there exits relationship between number of duplication and letters in the latter problem, the latter one is more difficult to solve compared to the former problem and Pisa and Metacat are unable to handle it.

\section{Conclusions}
\label{sec:conclusion}

In this paper, we proposed \textbf{N}eural l\textbf{o}gic \textbf{a}nalogy lear\textbf{n}ing (Noan), which is a dynamic neural architecture driven by differentiable logic reasoning to solve analogy problems.
In particular, each analogy problem is converted into logical expressions consisting of logical variables and basic logical operations (AND, OR, and NOT). Noan learns the logical variables as vector embeddings and learns each logical operation as a neural module. In this way, the integration of neural network and logical reasoning enables the model to capture the internal logical structure of the input letter strings. Then, the analogy learning problem becomes a True/False evaluation problem of the logical expressions. Experiments show that our machine learning-based Noan approach performs well on standard letter-string analogy datasets.

\bibliographystyle{iclr2022_conference}
\bibliography{reference}

\begin{thebibliography}{27}
\providecommand{\natexlab}[1]{#1}
\providecommand{\url}[1]{\texttt{#1}}
\expandafter\ifx\csname urlstyle\endcsname\relax
  \providecommand{\doi}[1]{doi: #1}\else
  \providecommand{\doi}{doi: \begingroup \urlstyle{rm}\Url}\fi

\bibitem[Barnden(1994)]{barnden1994connectionist}
John~A Barnden.
\newblock On the connectionist implementation of analogy and working memory
  matching.
\newblock 1994.

\bibitem[Bartha(2013)]{bartha2013analogy}
Paul Bartha.
\newblock Analogy and analogical reasoning.
\newblock 2013.

\bibitem[Chen et~al.(2021)Chen, Shi, Li, and Zhang]{chen2021neural}
Hanxiong Chen, Shaoyun Shi, Yunqi Li, and Yongfeng Zhang.
\newblock Neural collaborative reasoning.
\newblock In \emph{Proceedings of the Web Conference 2021}, pp.\  1516--1527,
  2021.

\bibitem[Chen et~al.(2022)Chen, Li, Shi, Liu, Zhu, and Zhang]{chen2022graph}
Hanxiong Chen, Yunqi Li, Shaoyun Shi, Shuchang Liu, He~Zhu, and Yongfeng Zhang.
\newblock Graph collaborative reasoning.
\newblock In \emph{Proceedings of the Fifteenth ACM International Conference on
  Web Search and Data Mining}, pp.\  75--84, 2022.

\bibitem[Falkenhainer et~al.(1989)Falkenhainer, Forbus, and
  Gentner]{falkenhainer1989structure}
Brian Falkenhainer, Kenneth~D Forbus, and Dedre Gentner.
\newblock The structure-mapping engine: Algorithm and examples.
\newblock \emph{Artificial intelligence}, 41\penalty0 (1):\penalty0 1--63,
  1989.

\bibitem[Finlayson \& Winston(2005)Finlayson and
  Winston]{finlayson2005intermediate}
Mark~Alan Finlayson and Patrick~Henry Winston.
\newblock Intermediate features and informational-level constraint on
  analogical retrieval.
\newblock In \emph{Proceedings of the Annual Meeting of the Cognitive Science
  Society}, volume~27, 2005.

\bibitem[Gentner(1983)]{gentner1983structure}
Dedre Gentner.
\newblock Structure-mapping: A theoretical framework for analogy.
\newblock \emph{Cognitive science}, 7\penalty0 (2):\penalty0 155--170, 1983.

\bibitem[Gentner \& Forbus(2011)Gentner and Forbus]{gentner2011computational}
Dedre Gentner and Kenneth~D Forbus.
\newblock Computational models of analogy.
\newblock \emph{Wiley interdisciplinary reviews: cognitive science}, 2\penalty0
  (3):\penalty0 266--276, 2011.

\bibitem[Gentner \& Smith(2012)Gentner and Smith]{gentner2012analogical}
Dedre Gentner and L~Smith.
\newblock Analogical reasoning.
\newblock In \emph{Encyclopedia of Human Behavior: Second Edition}, pp.\
  130--136. Elsevier Inc., 2012.

\bibitem[Greiner(1988)]{greiner1988learning}
Russell Greiner.
\newblock Learning by understanding analogies.
\newblock \emph{Artificial Intelligence}, 35\penalty0 (1):\penalty0 81--125,
  1988.

\bibitem[Gust et~al.(2006)Gust, K{\"u}hnberger, and Schmid]{gust2006metaphors}
Helmar Gust, Kai-Uwe K{\"u}hnberger, and Ute Schmid.
\newblock Metaphors and heuristic-driven theory projection (hdtp).
\newblock \emph{Theoretical Computer Science}, 354\penalty0 (1):\penalty0
  98--117, 2006.

\bibitem[Hofstadter(2001)]{hofstadter2001analogy}
Douglas~R Hofstadter.
\newblock Analogy as the core of cognition.
\newblock \emph{The analogical mind: Perspectives from cognitive science}, pp.\
   499--538, 2001.

\bibitem[Hofstadter \& Mitchell(1994)Hofstadter and
  Mitchell]{hofstadter1994copycat}
Douglas~R Hofstadter and Melanie Mitchell.
\newblock The copycat project: A model of mental fluidity and analogy-making.
\newblock 1994.

\bibitem[Holyoak \& Thagard(1989)Holyoak and Thagard]{holyoak1989computational}
Keith~J Holyoak and Paul~R Thagard.
\newblock A computational model of analogical problem solving.
\newblock \emph{Similarity and analogical reasoning}, 242266, 1989.

\bibitem[Holyoak et~al.(1995)Holyoak, Holyoak, and Thagard]{holyoak1995mental}
Keith~J Holyoak, Keith~James Holyoak, and Paul Thagard.
\newblock \emph{Mental leaps: Analogy in creative thought}.
\newblock MIT press, 1995.

\bibitem[Hummel \& Holyoak(1997)Hummel and Holyoak]{hummel1997distributed}
John~E Hummel and Keith~J Holyoak.
\newblock Distributed representations of structure: A theory of analogical
  access and mapping.
\newblock \emph{Psychological review}, 104\penalty0 (3):\penalty0 427, 1997.

\bibitem[Keane(1995)]{keane1995order}
Mark~T Keane.
\newblock On order effects in analogical mapping: Predicting human error using
  iam.
\newblock Technical report, Trinity College Dublin, Department of Computer
  Science, 1995.

\bibitem[Kokinov(1994)]{kokinov1994hybrid}
Boicho Kokinov.
\newblock A hybrid model of reasoning by analogy.
\newblock \emph{Advances in connectionist and neural computation theory},
  2:\penalty0 247--318, 1994.

\bibitem[Kokinov \& Petrov(2001)Kokinov and Petrov]{kokinov2001integrating}
Boicho Kokinov and Alexander Petrov.
\newblock Integrating memory and reasoning in analogy-making: The ambr model.
\newblock \emph{The analogical mind: Perspectives from cognitive science}, pp.\
   59--124, 2001.

\bibitem[Larkey \& Love(2003)Larkey and Love]{larkey2003cab}
Levi~B Larkey and Bradley~C Love.
\newblock Cab: Connectionist analogy builder.
\newblock \emph{Cognitive Science}, 27\penalty0 (5):\penalty0 781--794, 2003.

\bibitem[Li et~al.(2008)Li, Vit{\'a}nyi, et~al.]{li2008introduction}
Ming Li, Paul Vit{\'a}nyi, et~al.
\newblock \emph{An introduction to Kolmogorov complexity and its applications},
  volume~3.
\newblock Springer, 2008.

\bibitem[Marshall \& Hofstadter(1997)Marshall and
  Hofstadter]{marshall1997metacat}
James~B Marshall and Douglas~R Hofstadter.
\newblock The metacat project: a self-watching model of analogy-making.
\newblock \emph{Cognitive Studies: Bulletin of the Japanese Cognitive Science
  Society}, 4\penalty0 (4):\penalty0 4\_57--4\_71, 1997.

\bibitem[Mikolov et~al.(2013)Mikolov, Sutskever, Chen, Corrado, and
  Dean]{mikolov2013distributed}
Tomas Mikolov, Ilya Sutskever, Kai Chen, Greg Corrado, and Jeffrey Dean.
\newblock Distributed representations of words and phrases and their
  compositionality.
\newblock \emph{arXiv preprint arXiv:1310.4546}, 2013.

\bibitem[Murena et~al.(2017)Murena, Dessalles, and
  Cornu{\'e}jols]{murena2017complexity}
Pierre-Alexandre Murena, Jean-Louis Dessalles, and Antoine Cornu{\'e}jols.
\newblock A complexity based approach for solving hofstadter's analogies.
\newblock In \emph{ICCBR (Workshops)}, 2017.

\bibitem[Rijsdijk \& Sileno()Rijsdijk and Sileno]{rijsdijksolving}
Geerten Rijsdijk and Giovanni Sileno.
\newblock Solving hofstadter’s analogies using structural information theory.

\bibitem[Shi et~al.(2020)Shi, Chen, Ma, Mao, Zhang, and Zhang]{shi2020neural}
Shaoyun Shi, Hanxiong Chen, Weizhi Ma, Jiaxin Mao, Min Zhang, and Yongfeng
  Zhang.
\newblock Neural logic reasoning.
\newblock In \emph{Proceedings of the 29th ACM International Conference on
  Information \& Knowledge Management}, pp.\  1365--1374, 2020.

\bibitem[Thagard et~al.(1990)Thagard, Holyoak, Nelson, and
  Gochfeld]{thagard1990analog}
Paul Thagard, Keith~J Holyoak, Greg Nelson, and David Gochfeld.
\newblock Analog retrieval by constraint satisfaction.
\newblock \emph{Artificial intelligence}, 46\penalty0 (3):\penalty0 259--310,
  1990.

\end{thebibliography}

\clearpage

\appendix

\section{Supplemental Materials}

\subsection{Related Work}
\label{sec:related_work}

Analogy is a core cognition of human beings \citep{hofstadter2001analogy}. This is because analogy is representative of human thinking that is structure flexible and sensitive \citep{barnden1994connectionist}, and analogy is a mental tool that is ubiquitously used in human reasoning \citep{holyoak1995mental}. To understand human analogy, some theories were proposed by cognitive psychologist.
\citeauthor{gentner1983structure} proposed a structure mapping theory for analogy including relations between objects are mapped from base to target and the systematicity defines the particular relations \citep{gentner1983structure}. 
\citeauthor{hummel1997distributed} proposed a theory of analogical access and mapping which simultaneously achieves the flexibility of a connectionist system and the structure sensitivity of a symbolic system \citep{hummel1997distributed}.

Also, many general computational analogy algorithms were developed to help people study the main analogy processes such as analog retrieval and similarity structure mapping, where retrieval is to find an analog that is similar to it with a given situation while mapping is to align two given situations structurally to produce a set of correspondences \citep{gentner2011computational}. Almost all models aim to capture mapping structures in analogies, such as ACME \citep{holyoak1989computational}, AMBR \citep{kokinov2001integrating}, CAB \citep{larkey2003cab}, HDTP \citep{gust2006metaphors}, IAM \citep{keane1995order}, NLAG \citep{greiner1988learning}, SME \citep{falkenhainer1989structure}, and Winston \citep{finlayson2005intermediate}.  
Besides, ARCS \citep{thagard1990analog} focuses on both retrieval and mapping processes, DUAL \citep{kokinov1994hybrid} is engaged in the processes including encoding, retrieval and mapping.

One typical system is CopyCat \citep{hofstadter1994copycat}. The goal of Copycat is to take concepts and understand the flexible perception and analogy-making of human beings through solving letter-string analogy problems. 
CopyCat was later updated to MetaCat \citep{marshall1997metacat} which can store different answers in memory and continue to search for alternative answers. 

It is supposed that more concepts are needed to solve more complex analogy problems, however, CopyCat and MetaCat cannot consider as many concepts as human beings.
One alternative is complexity-based approach. For example, \cite{murena2017complexity} proposed an complexity based approach to solving letter-string analogies. To describe analogy problems, basic rules for a new generative language were proposed. 
With the language, the Kolmogorov complexity can be used to measure the relevance in analogical reasoning. Then, an analogy problem can be solved by taking the solution with the minimal complexity. 
Similar to the CopyCat and the MetaCat, the Pisa \citep{rijsdijksolving} algorithm is based on the idea that a certain structure between initial string and modified string exists and can be adopted to the target string.
It first extracts structures between initial string and modified string by compressing two strings and applying Structural Information Theory (SIT) which proposes to apply simplicity principle to find an encoding of a string with minimal complexity.

\subsection{Preliminaries and Problem Formalization}
\label{sec:preliminary}

In this section, we will present a brief introduction about applying logical operators and basic logic laws to the analogy solving. Typically, there are three fundamental operations: AND (conjunction), OR (disjunction), and NOT (negation). In logical reasoning, each variable $x$ represents a $literal$. A clause is literals with a flat operation, such as $x\land y$. An expression is clauses with operations, such as $(x\land y)\lor (a\land b \land c)$. We follow universal laws in propositional logic about NOT, AND, and OR. Another important law in this paper is the De Morgan's Law, which can can expressed as:
$$\lnot (x \land y) \Longleftrightarrow  \lnot x \lor \lnot y,~~~\lnot (x \lor y) \Longleftrightarrow  \lnot x \land \lnot y$$
We also need to introduce another secondary logical operation $x\rightarrow y$, which is also known as material  implication. This operation states a logical equivalence which could be formulated as:
$$x\rightarrow y \Longleftrightarrow \lnot x \lor y$$
Although the above propositional logic knowledge can help convert natural analogies into symbolic reasoning, it fails to accomplish continuous optimization because of its lack of ability to learn from given data. So we adopt the idea of distributed representation learning \citep{mikolov2013distributed} and then build a neural-symbolic framework in a continuous manner. In this framework, each literal $x$ represents a character, and is transformed as an embedding vector \textup{x}. And each logical operation, such as AND, OR, NOT, is transformed as a neural module, e.g., AND(\textbf{x}, \textbf{y}). In this way, each expression can be transformed as a neural architecture that have the ability of making True/False judgement.

\subsection{Tables and Figures}

\begin{table*}[h!]
  \begin{center}
    \caption{Logical regularizers and the corresponding logical rules}
    \label{tab:logicalrules}
    \begin{small}
    \begin{tabular}{l l l l }
    \hline
            & Logical Rule    & Equation    & Logic Regularizer $r_i$ \\
    \hline

    NOT     &Negation         &$\lnot T = F$         &$r_1 = \sum_{w\in W \cup \{T\}} Sim(\rm{NOT}(\textbf{w}),\textbf{w})$   \\
            &Double Negation  &$\lnot(\lnot w) = w$  &$r_2 = \sum_{w\in W} 1-Sim(\rm{NOT}(\rm{NOT}(\textbf{w})),\textbf{w})$   \\   
    \hline
            &Identify         &$w \land T = w$       &$r_3 = \sum_{w\in W} 1-Sim(\rm{AND}(\textbf{w}, \textbf{T}),\textbf{w})$   \\
    AND     &Annihilator      &$w \land F = F$       &$r_4 = \sum_{w\in W} 1-Sim(\rm{AND}(\textbf{w}, \textbf{F}),\textbf{F})$   \\
            &Idempotence      &$w \land w = w$       &$r_5 = \sum_{w\in W} 1-Sim(\rm{AND}(\textbf{w}, \textbf{w}),\textbf{w})$   \\
            &Complementation  &$w \land \lnot w = F$ &$r_6 = \sum_{w\in W} 1-Sim(\rm{AND}(\textbf{w}, \rm{NOT}(\textbf{w})),\textbf{F})$   \\
    \hline
            &Identify         &$w \lor F = w$        &$r_7 = \sum_{w\in W} 1-Sim(\rm{OR}(\textbf{w}, \textbf{F}),\textbf{w})$   \\
    OR      &Annihilator      &$w \lor T = T$        &$r_8 = \sum_{w\in W} 1-Sim(\rm{OR}(\textbf{w}, \textbf{T}),\textbf{T})$   \\
            &Idempotence      &$w \lor w = w$        &$r_9 = \sum_{w\in W} 1-Sim(\rm{OR}(\textbf{w}, \textbf{w}),\textbf{w})$   \\
            &Complementation  &$w \lor \lnot = T$    &$r_{10} = \sum_{w\in W} 1-Sim(\rm{OR}(\textbf{w}, \rm{NOT}(\textbf{w})),\textbf{T})$   \\
            
    \hline
    \end{tabular}
    \end{small}
  \end{center}
\end{table*}

\begin{table}[h!]
  \begin{center}
    \caption{Human answers to analogies of form ABC:ABD::X:? from Murena's dataset, along with at which position the same answers were given by Noan ($P_n$), Pisa ($P_p$) and Metacat ($P_m$)}
    \label{tab:dataset1}
    \begin{tabular}{|c |c |c |c |c |c |}
    \hline
    Given X  & Solutions   & Selected    & $P_n$    & $P_p$  & $P_m$ \\
    \hline

    IJK     &IJL     &93$\%$   &1  &1  &1 \\
            &IJD     &2.9$\%$  &2  &$\infty$  &$\infty$ \\   
    \hline
     BCA     &BCB     &49$\%$   &1  &3  &2 \\
            &BDA     &43$\%$   &2  &1  &1 \\
    \hline
    AABABC  &AABABD  &74$\%$   &1  &1  &1 \\
            &AACABD  &12$\%$   &2  &$\infty$  &$\infty$ \\
    \hline
    IJKLM   &IJKLN   &62$\%$   &1  &1  &1 \\
            &IJLLM   &15$\%$   &2  &$\infty$  &$\infty$ \\
    \hline
    KJI     &KJJ     &37$\%$   &1  &1  &1 \\
            &LJI     &32$\%$   &2  &$\infty$  &2 \\
    \hline
    ACE     &ACF     &63$\%$   &1  &1  &1 \\
            &ACG     &8.9$\%$  &7  &$\infty$  &$\infty$ \\
    \hline
    BCD     &BCE     &81$\%$   &2  &2  &2 \\
            &BDE     &5.9$\%$  &1  &1  &1 \\
    \hline
    \rowcolor{green} IJJKKK  &IJJLLL  &40$\%$   &1  &1  &1 \\
    \rowcolor{green}        &IJJKKL  &25$\%$   &2  &2  &2 \\ 
    \hline
    XYZ     &XYA     &85$\%$   &1  &1  &1 \\
            &IJD     &4.4$\%$  &11 &$\infty$  &$\infty$ \\
    \hline
    \rowcolor{green} RSSTTT  &RSSUUU  &41$\%$   &1  &1  &1 \\
    \rowcolor{green}        &RSSTTU  &31$\%$   &2  &2  &$\infty$ \\
    \hline
    \rowcolor{red} MRRJJJ  &MRRJJK  &28$\%$   &2  &2  &1 \\
    \rowcolor{red}        &MRRKKK  &19$\%$   &1  &1  &2 \\

    \hline
    \end{tabular}
  \end{center}
\end{table}

\begin{table}[h!]
  \begin{center}
    \caption{Human answers to analogies from Rijsdijk's dataset, along with at which position the same answers were given by Noan ($P_n$), Pisa ($P_p$) and Metacat ($P_m$)}
    \label{tab:dataset2}
    \small
    \setlength{\tabcolsep}{0pt}{
    \begin{tabular}{|c |c |c |c |c |c |}
    \hline
    Given problem  & Solutions   & Selected   & $P_n$  & $P_p$  & $P_m$ \\
    \hline
    ABA:ACA::               &AEA              &97.1$\%$   &1  &1  &1  \\
    ADA:?                   &AFA              &2.9$\%$    &2  &$\infty$  &$\infty$  \\   
    \hline
    \rowcolor{red} ABAC:ADAE::             &DAEA             &60$\%$     &2  &2  &$\infty$  \\
    \rowcolor{red} BACA:?                  &BCCC             &28.6$\%$   &1  &21 &$\infty$  \\
    \hline
    AE:BD::                 &DB               &68.5$\%$   &1  &3  &1  \\
    CC:?                    &CC               &17.1$\%$   &2  &$\infty$  &2  \\
    \hline
    ABBB:AAAB::             &IIJJJ            &57.1$\%$   &1  &1  &$\infty$  \\
    IIIJJ:?                 &JJIII            &14.3$\%$   &2  &$\infty$  &$\infty$  \\
    \hline
    ABC:CBA::               &IJKLM            &88.6$\%$   &1  &1  &1  \\
    MLKJI:?                 &-                &-          &$\infty$  &$\infty$  &$\infty$  \\
    \hline
    ABCB:ABCB::             &Q                &100$\%$    &1  &1  &$\infty$  \\
    Q:?                     &-                &-          &$\infty$  &$\infty$  &$\infty$  \\
    \hline
    \rowcolor{red} ABC:BAC::               &JIKL             &54.3$\%$   &2  &$\infty$  &$\infty$  \\
    \rowcolor{red} IJKL:?                  &KIJL             &14.3$\%$   &3  &2  &$\infty$  \\
    \hline
    ABACA:BC::              &AA               &57.1$\%$   &1  &1  &$\infty$  \\
    BACAD:?                 &BCD              &31.4$\%$   &3  &$\infty$  &$\infty$  \\ 
    \hline
    AB:ABC::                &IJKLM            &85.7$\%$   &1  &1  &1  \\
    IJKL:?                  &IJKLMN           &11.4$\%$   &2  &$\infty$  &$\infty$  \\
    \hline
    ABC:ABBACCC::           &FEEFDDD          &91.4$\%$   &1  &2  &1  \\
    FED:?                   &-                &-          &$\infty$  &$\infty$  &$\infty$  \\
    \hline
    \rowcolor{green} ABC:BBC::               &JKM              &57.1$\%$   &1  &7  &$\infty$  \\
    \rowcolor{green} IKM:?                   &KKM              &37.1$\%$   &2  &2  &$\infty$  \\
    \hline
    \rowcolor{green} ABAC:ACAB::             &DGFE             &68.6$\%$   &1  &2  &$\infty$  \\
    \rowcolor{green} DEFG:?                  &FGDE             &14.3$\%$   &2  &1  &$\infty$  \\
    \hline
    ABC:ABD::               &DBA              &51.4$\%$   &1  &1  &2  \\
    CBA:?                   &CBB              &45.7$\%$   &2  &2  &1  \\
    \hline
    ABAC:ADAE::             &FDFE             &94.3$\%$   &1  &1  &$\infty$  \\
    FBFC:?                  &FDFA             &2.9$\%$    &6  &$\infty$  &$\infty$  \\
    \hline
    \rowcolor{green} ABCD:CDAB::             &LMNIJK           &80.0$\%$   &1  &$\infty$  &$\infty$  \\
    \rowcolor{green} IJKLMN:?                &-                &-          &$\infty$  &$\infty$  &$\infty$  \\
    \hline
    {\scriptsize ABC:AAABBBCCC::}         & {\scriptsize AAABBBCCCDDD}     &74.3$\%$   &1  &1  &1  \\
    {\scriptsize ABCD:?}                  & {\scriptsize AAAABBBBCCCCDDDD} &17.1$\%$   &2  &$\infty$  &$\infty$  \\
    \hline
    \rowcolor{green} ABC:ABBCCC::            &ABBCCCDDDD       &85.7$\%$   &1  &$\infty$  &$\infty$  \\
    \rowcolor{green} ABCD:?                  &ABBCCCDDD        &8.6$\%$    &2  &1  &$\infty$  \\
    \hline
    ABBCCC:DDDEEF::         &DEEFFF           &77.1$\%$   &1  &1  &$\infty$  \\
    AAABBC:?                &DCCDDF           &8.6$\%$    &3  &$\infty$  &$\infty$  \\
    \hline
    A:AA::                  &AAAAAA           &62.8$\%$   &1  &1  &$\infty$  \\
    AAA:?                   &AAAA             &25.7$\%$   &2  &2  &1  \\
    \hline
    \rowcolor{green} ABBA:BAAB::             &JILK             &71.4$\%$   &1  &$\infty$  &$\infty$  \\
    \rowcolor{green} IJKL:?                  &JIJM             &11.4$\%$   &2  &5  &$\infty$  \\

    \hline
    \end{tabular}}
  \end{center}
\end{table}

\end{document}